\documentclass[12pt]{article}
\usepackage{times}
\usepackage{graphicx}
\usepackage[square,sort,comma,numbers]{natbib}
\usepackage{amsmath}
\usepackage{epstopdf}

\title{Performance comparison of State-of-the-art Missing Value Imputation Algorithms on Some Bench mark Datasets}
 \author{M.~Naresh Kumar, \\
  \multicolumn{1}{p{.7\textwidth}}{\centering\emph{National Remote Sensing Centre (ISRO),
  India, nareshkumar\_m@nrsc.gov.in}}}

\begin{document}
\maketitle

\begin{abstract}
Decision making from data involves identifying a set of attributes that contribute to effective decision making through computational intelligence. The presence of missing values greatly influences the selection of right set of attributes and this renders degradation in classification accuracies of the classifiers. As missing values are quite common in data collection phase during field experiments or clinical trails appropriate handling would improve the classifier performance. In this paper we present a review of recently developed missing value imputation algorithms and compare their performance on some bench mark datasets. 
\end{abstract}

\newpage
\section{Introduction}
The presence of missing values influences the selection of appropriate set of attributes that render degradation in classification accuracies of the classifiers. Missing values are a common problem in almost all real world data sets~\citep{1} used in knowledge discovery and data mining(KDD) applications. Specifically they are more frequent in clinical databases~\citep{ColleenMNorris2000, Geert2007, Cios2002} and temporal climate databases~\citep{Glasbey1995, Antti2010}. Their presence would greatly affect the performance of classifiers~\citep{Farhangfar2008}.

The missing values in the databases may arise due various reasons such as the value being lost (erased or deleted) or not recorded, incorrect measurements, equipment errors, or possibly due to an expert not attaching any importance to a particular procedure. The incomplete data can be identified by looking for null values in the data set. However, this is not always true, since missing values can appear in the form of outliers or even wrong data (i.e. out of boundaries)~\citep{Pearson2005}.

There are numerous methods for predicting or approximating missing values. Single imputation strategies involve using the mean, median or mode~\citep{Schafer1997} or regression-based methods~\citep{HortonNJ2001} to impute the missing values. Traditional approaches of handling missing values like complete case analysis, overall mean imputation and missing-indicator method~\citep{Heijden2006} can lead to biased estimates and may either reduce or exaggerate the statistical power. Each of these distortions can lead to invalid conclusions. Statistical methods of handling missing values depends on the application of maximum likelihood and expectation maximization algorithms~\citep{Allison2002, RoderickJALittle2002, Schafer1997}. Some of these methods would work only for certain types of attributes either nominal or numeric. Machine learning approaches like neural networks with genetic algorithms~\citep{MussaABDELLA2006}, neural networks with particle swarm optimization~\citep{Qiao2005} have been used to approximate the missing values.  The use of neural networks comes with a greater cost in terms of computation and training. Methods like radial basis function networks, support vector machines and principal component analysis have been studied for estimating the missing values. A new index and a distance measure was developed in~\citep{rao2012new,hari2013novel} for imputing the missing values which have resulted in improved classification accuracies. For an extensive treatment of these methods we refer the readers to Marwala~\citep{Marwala2009}.

The following are the main objectives of the present work:
 \begin{itemize}
    \item[(i)] a review on imputation algorithm (RNI-II) developed by Rao and Naresh~\citep{rao2012new,hari2013novel};
    \item[(ii)] to consider the case in which the attribute values are missing randomly in real world data sets;
    \item[(iii)] to compare the performance of RNI-II algorithm with other well known imputation methods in improving the classification accuracies of the decision tree classifiers such as C4.5~\citep{Quinlan1993} and Genetic Algorithm-C4.5~\citep{CF2004};
		
	\end{itemize}

These claims are realized through standard databases available University of California Irvin (UCI) machine learning and the Keel data repositories~\citep{Frank2010, Alcala-Fdez2010}.

This paper is organized as follows: Section~\ref{chap9:sec2} describes the preliminaries of different missing value handling methods. A review on some recently developed missing value imputation algorithms (RNI-II) forms the core of Section~\ref{sec3}. The experiments on some benchmark datasets are described and the results are discussed in Section~\ref{chap9:sec5}. Conclusions and discussion of the results are present in Section~\ref{chap9:sec6}.

\section{Imputation Algorithms- A Review}
\label{chap9:sec2}
The following are the main imputation algorithms popularly employed in handling missing values in data sets :
 
\begin{itemize}
  \item Imputation with K-Nearest Neighbour (KNNI)~\citep{BM2003}: the algorithm computes the k-nearest neighbour for each of the missing values and imputes a value from them. For nominal values, the most common value among all neighbours is taken, and for numerical values the mean value is computed. To obtain the proximity the Euclidean distance (a case of an Lp norm distance) norm is considered;

\item Weighted Imputation with K-Nearest Neighbour (WKNNI)~\citep{TCSBHTBA2001}: the weighted k-nearest neighbour method selects the instances with similar values (in terms of distance) to the record having the MV. The estimated value takes into account the distances among the neighbours, using a weighted mean or the most repeated value according to the distance;

 \item K-Means clustering Imputation (KMI)~\citep{DSSL2004}: in K-means clustering imputation, the intra-cluster dissimilarity is measured by the addition of distances among the objects and the centroid of the cluster which they are assigned to. A cluster centroid represents the mean value of the objects in the cluster. Once the clusters have converged, the last process in KMI is to fill in all the non-reference attributes for each incomplete object based on the cluster information. Data objects that belong to the same cluster are taken as nearest neighbours of each other, and a nearest neighbour algorithm is employed to replace missing data;

\item Fuzzy K-Means clustering based Imputation (FKMI)~\citep{AR2004, DSSL2004}: in fuzzy clustering, each data object $x_i$ has a membership function which describes the degree which this data object belongs to a certain cluster $v_k$.  In this process, the data object does not belong to a concrete cluster represented by a cluster centroid (as done in the basic k-mean clustering algorithm), because each data object belongs to all $k$ clusters with different membership degrees. The non-reference attributes for each incomplete data object $x_i$ based on the information about membership degrees and the values of cluster centroid;

\item Support Vector Machine  Imputation (SVMI)~\citep{FCYYC2005}: is a Support Vector Machine (SVM) regression based algorithm to fill in missing data, by setting the decision attributes (output or classes) as the condition attributes (input attributes) and the condition attributes as the decision attributes.Subsequently SVM regression is used to predict the missing condition attribute values. In the first step the values without missing values are considered. In the next step the conditional attributes (input attributes), some of those values among them are missing, as the decision attribute (output attribute), and the decision attributes as the condition attributes by contraries. Finally, SVM regression is used to predict the decision attribute values.
\end{itemize}

\section{Imputation Algorithm (RNI-II) proposed in ~\citep{hari2013novel}}
\label{sec3}
The authors in ~\citep{rao2012new} have developed a mean imputation procedure based on a novel indexing measure to identify the best record that fits the missing value for imputation. As a further improvement the authors in~\citep{hari2013novel} develop a imputation procedure considering the higher order statistics concerning the attributes with the following salient features:
\begin{itemize}
    \item considers incomplete data sets as input and returns the completed data sets with all instances of missing attribute values filled with possible values;
    \item handles data sets with attributes of different types for example nominal or numeric;
    \item last attribute in the data set must be a decision attribute of the type nominal and may belong to multiple classes;
    \item the values to be imputed are obtained by using the data from within the data set;
    \item does not require additional parameters such as $k$ in KNN imputation.
\end{itemize}

The most popular metric for quantifying the similarity between the two records is the Euclidean distance. Even though this metric is simpler to compute, it suffers from the following drawbacks~\citep{Tadashi2009}:
\begin{itemize}
\item Sensitive to the scales of the features involved;
\item It does not account for correlation between the features.
\end{itemize}

With this motivation the authors in~\citep{hari2013novel} define a new indexing measure that accounts for the interaction between the features and their distribution. Based on the index we then compute the distances between the tuples.

\subsection{Formalization of the distance measure}
\label{distformal}
For the sake of completeness we present the measure in this section. For more details we refer our readers to~\citep{hari2013novel}.

Let \emph{S} denote the collection of all data records, represented $A_{ij}$ for $\emph{i}=1,2,\hdots,m$ and $\emph{j}=1,2,\hdots,n-1$. The attributes can either be a nominal (categorical) or numeric type. Our objective here is to define a distance measure for all pairs of the elements of \emph{S}. 

In the first case the decision of the two tuples may be same ~ $A_{in}=A_{kn}$. Let \emph{A} denote the collection of all members of \emph{S} that belong to the same decision class to which $R_{i}$ and $R_{k}$ belong. If the attribute to which the column $C_{l}$ belongs is nominal then the set $A$ can be written as a union of subsets of distinctive elements or collections of the attributes. That is $\emph{A}=B_{\gamma_{p_{1l}}}\bigcup B_{\gamma_{p_{2l}}}\bigcup, \ldots,\bigcup B_{\gamma_{p_{sl}}}$. The  index for column $C_{l}$ of S for two tuples $R_{i}$,$R_{k}$ may be computed as 
  \begin{eqnarray*}
    I_{C_{l}}(R_{i},R_{k})=\left\{
                             \begin{array}{ll}
                              \min\{\frac{\gamma_{p_{il}}}{\gamma},\frac{\gamma_{q_{kl}}}{\gamma}\}, & \hbox{for $i \neq k$;} \\
                               0, & \hbox{otherwise,}
                             \end{array}
                           \right.
\end{eqnarray*}
\noindent  where $\gamma_{p_{il}}$ represents the cardinality of the subset $B_{\gamma_{p_{il}}}$, all of whose elements have first co-ordinates $A_{il}$ , $\gamma_{q_{kl}}$ represents the cardinality of that subset $B_{\gamma_{q_{kl}}}$, all of whose elements have first co-ordinates $A_{kl}$ and $\gamma=\gamma_{p_{1l}}+ \gamma_{p_{2l}}+ \ldots,+\gamma_{p_{sl}}$ represents the cardinality of the set $A$.

If the attribute $C_{l}$ is of numeric type and for fractional numbers the set $P_{l}$ for $l=1,2,\dots,n-1$ may be considered as a collection of all the members of the column $C_{l}$. Based on the skewness of the dataset the index is computed as 

\begin{eqnarray*}
    I_{C_{l}}(R_{i},R_{k})=\left\{
                             \begin{array}{ll}
                             \min\{\frac{\tau_{l}}{\gamma},\frac{\rho_{l}}{\gamma}\}, & \hbox{for $i \neq k$;} \\
                               0, & \hbox{otherwise,}
                             \end{array}
                           \right.
\end{eqnarray*}
In the above definition $\gamma$ represents the cardinality of the set $P_{l}$ and $\tau_{l}$ and $\rho_{l}$ be the cardinalities of the sets $M_{l}$ and $N_{l}$ where in $M$ and $N$ are subsets constructed out of the subset of elements belonging to skewness being less than or greater than zero.

The decision attribute of the tuples may be different i.e ~ $A_{in}\neq A_{kn}$. Under this condition 
clearly $R_{i}$ and $R_{k}$ belong to two different decision classes. Consider the subsets $P_{i}$ and $Q_{k}$ consisting of members of $\emph{S}$ that share the same decision with $R_{i}$ and $R_{k}$ respectively. Clearly $P_{i}\bigcap Q_{k} = \emptyset$. The members of the column $C_{l}$ of S i.e $(A_{1l},A_{2l},\ldots,A_{ml})^{T}$ are of nominal or categorical type then the indexing measure between the two records $R_{i}$ and $R_{k}$  is defined as
\begin{eqnarray*}
    I_{C_{l}}(R_{i},R_{k})=\left\{
                             \begin{array}{ll}
                               \max\{\frac{\beta_{rl}}{\delta_{sl}+\beta_{rl}},\frac{\delta_{sl}}{\delta_{sl}+\beta_{rl}}\}, & \hbox{for $i \neq k$;} \\
                               0, & \hbox{otherwise,}
                             \end{array}
                           \right.
\end{eqnarray*}

where $\beta_{rl}$ represents the cardinality of the subset $P_{\beta_{rl}}$ all of whose elements have first co-ordinates $A_{il}$ in set $P_{l}$ and $\delta_{sl}$ represents the cardinality of that subset $Q_{\delta_{sl}}$, all of whose elements have first co-ordinates $A_{kl}$ in set $Q_{l}$. The set $P$ and $Q$ are sets having elements matching with the decision attribute of $R_i$ and $R_k $\label{case2:item1}
If the members of the column $C_{l}$ of S i.e $(A_{1l},A_{2l},\ldots,A_{ml})^{T}$   are of numeric type and for fractional numbers the index $I_{C_{l}}(R_{i},R_{k})$ between the two records $R_{i},R_{k}$ is defined as
\begin{eqnarray*}
    I_{C_{l}}(R_{i},R_{k})=\left\{
                               \begin{array}{ll}
                             \min\{\frac{\beta_{l}}{\lambda},\frac{\delta_{l}}{\lambda}\}, & \hbox{for $i \neq k$;} \\
                               0, & \hbox{otherwise,}
                             \end{array}
                           \right.
\end{eqnarray*}
In the above definition $\beta_{l}$ and $\delta_{l}$ represents the cardinalities of the sets $R_{l}$ and $S_{l}$ respectively where $R$ and $S$ denote the sets with skewness equal to zero and greater than zero. The sum of the cardinalities of the sets $P_{l}$ and $Q_{l}$ is represented by $\lambda$.

The elements of set $D$ consisting of the distances between the tuples in an ascending order is constructed. To identify the nearest tuples the score $\alpha(x_{k})=\frac{(x_{k}-median(x))}{median{|x_{i}-median(x)|}}$ where $\{x_{1},x_{2},\ldots,x_{n} \}$ denote the distances of $R$ from $R_{k}$ is defined. The data records in set $S$ whose distances from the record $R$ satisfies the condition $\alpha(x_{k}) \leq 0$ is collected and is designated as $P$. 
For nominal type the frequency of each categorical value of the categorical attribute is computed and the highest categorical value of the frequent item set is imputed. If the type of attribute is numeric and non-integer for each element in set $B$ compute the quantity $\beta(j)$=$\frac{1}{B(j)}$ $\forall j=1,\ldots,\gamma$ where $\gamma$ denote the cardinality of the set $B$ is computed. The weight matrix is computed as  $W(j)=\frac{\beta_{j}}{\sum_{i=1}^{j} \beta(j)}$ $\forall j=1,\ldots,\gamma$. The value to be imputed may be taken as $\sum_{i=1}^{j} P(j).W(j)$ $\forall j=1,\ldots,\gamma$.

\section{Experiments and Results}
\label{chap9:sec5}
\subsection{Description of the Data sets}
We have selected nine($9$) data sets having missing values ranging from $2.19\%$ to $70.58\%$ in different types of attributes(integer, real (fractional or non-integer) and categorical(nominal)) from the Keel and UCI machine learning data repositories~\citep{Alcala-Fdez2010,Frank2010}.

We have selected the data sets pertaining to different problem domains such as: diagnosis of mammographic lesion-Mammographic (MAM), breast cancer identification-Breast (BRE), predicting erythemato-squamous skin diseases- Dermatology (DER), detect the presence of heart disease in the patients-Cleveland (CLE), data records pertaining to patients suffering from hyperthyroidism -    Newthyroid (NTH), mitigating process delays due to rotogravure printing-Bands (BAN), classifying the U.S. House of Representatives Congressmen as Republican or Democrats based on the votes-House-votes (HOV), credit card applications approval- Australian (AUS), identification of different species of Iris plant-Iris (IRI). The details of the data sets are given in Table~\ref{chap9:tab3}, with summary of their properties. The column labeled as ``\% MV.'' indicates the percentage of the missing values in the data set. The column labeled as ``\% Ex.'' refers to the percentage of examples in the data set which have at least one missing value.
\begin{table}[!ht]
\caption{Data sets used in the experiments}
\label{chap9:tab3}
 \centering
\begin{tabular} {llllll}
\hline
\bfseries Data set & \bfseries \#Attr. & \bfseries \#Ex.& \bfseries \#CL& \bfseries \%MV&\bfseries \#DS(KB)\\
&\bfseries (R/I/N)&&&&\\
\hline \hline
Dermatology(DER) & 34  & 366 & 6 & 2.19&442\\
&(0/34/0)&&&&\\
Mammographic(MAM) & 5& 961 & 2 & 13.63 & 194\\
& (0/5/0) &&&&\\
Breast(BRE) & 9 & 286 &2 & 3.15 &241\\
&(0/0/9)&&&&\\
House-votes(HOV) & 16 & 435 & 2 & 46.67&303\\
&(0/0/16)&&&&\\
Cleveland(CLE) & 13 & 303 & 5 & 1.98&253 \\
&(13/0/0)&&&&\\
Iris(IRI)&4 &150&	3	&32.67 &52.1\\
&(4/0/0)	&&&&\\
Bands(BAN) & 19 & 539 & 2 & 32.28&649\\
& (13/6/0)&&&&\\
Newthyroid(NTH)&	5 (4/1/0)&	215	&3&	35.35 &70.4\\
& (4/1/0) &&&&\\
Australian(AUS) &14 &690&	2&	70.58&478 \\
&(3/5/6)&&&&\\
\end{tabular}
\end{table}

\subsection{Experimental Methodology}
\begin{figure}[!ht]
\centering
  \includegraphics[width=3.5in]{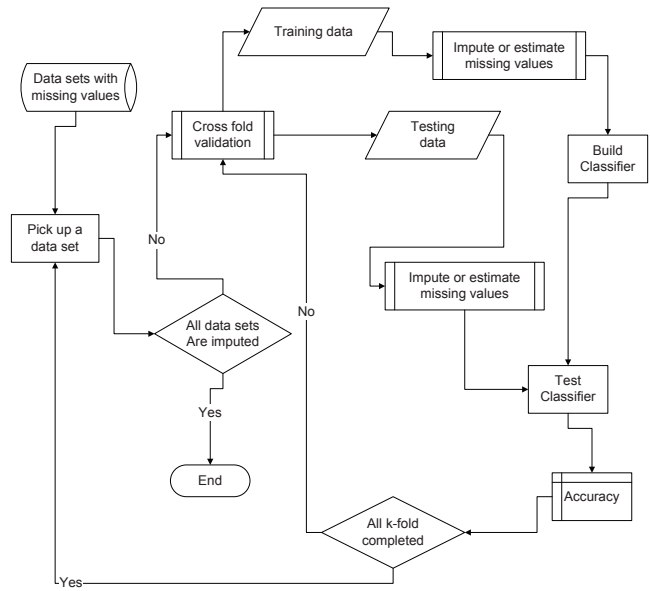}\\
  \caption{Structure of the proposed methodology}\label{chap9:fig7}
\end{figure}

 The experimental structure of our proposed methodology is shown in Figure~\ref{chap9:fig7}. The data sets are considered one at a time and each of the data set is divided into training and testing data sets using a stratified $k$ fold cross validation method~\citep{Littlestone1988} with $k=10$ . For each fold, the training data set is considered for imputation using methodologies and  the respective classifiers are built. The testing data sets that correspond to the training data sets are used in classifying the records. The average accuracy is computed by taking the average of the classes that are predicted correctly relative to each classifier. The same methodology is followed for all the data sets considered in our experiment.

A non-parametric statistical test proposed by Wilcoxon~\citep{Wilcoxon1945} is used to bench mark the performance of the RNI-II algorithm with respect to other imputation algorithms in terms of the accuracies obtained using C4.5 and GA-C4.5. Wilcoxon signed rank non-parametric test is a popular measure generally employed to compare the results across multiple data sets~\citep{Bouckaert2003, Dietterich1998, Demvsar2006, Garcia2008}.
\subsection{Parameters used in the study}
The parameters used by each imputation method is given in Table~\ref{chap9:tab1}. The values chosen are as recommended by the respective authors.
\begin{table}[!ht]
\caption {Methods and Parameters}
 \label {chap9:tab1}
 \centering
\begin{tabular}{cc}
\hline
\bfseries Method & \bfseries Parameters\\
 \hline \hline
{FKMI} & K=3 \\ & error=100  \\
& iterations=100 \\ & m=1.5\\
{KMI} & K=10 \\ & error=100  \\
& iterations=100 \\
{KNNI} & K=10 \\
{WKNNI} & K=10 \\
\end{tabular}
\end{table}

\subsection{Experimental comparison of RNI-II with other imputation methods }
In this section we present the analysis of our experimental results. The mean test accuracies of C4.5 and GA-C4.5 decision tree classifiers for all the $10$ folds of the cross-validation data sets are computed and the results obtained invoking are tabulated in Table~\ref{chap9:tab6} and Table~\ref{chap9:tab8} respectively.

From the Table~\ref{chap9:tab6} we observe that the RNI-II algorithm has improved the test accuracies in all the nine data sets for C4.5 when compared with other imputation methods like FKMI, KMI, KNNI and WKNNI. In case of the data sets BAN, BRE, DER and MAM the accuracies of C4.5 classifier obtained using the RNI-II algorithm is comparable with the SVMI algorithm. In case of accuracies obtained using GA-C4.5 classifier Table~\ref{chap9:tab8} the RNI-II algorithm outperformed FKMI, KMI, KNNI, WKNNI for data sets AUS, BAN, HOV, IRM, MAM, CLE and NTH and equalled the accuracies in data sets BRE and DER. Our RNI-II algorithm outperformed SVMI for data sets AUS, BAN, CLE and MAM.

\begin{table}[!ht]
   \centering
  \caption{C4.5 Classifier Test accuracies of decision tree classifiers}
  
  \label{chap9:tab6}
 \begin{tabular}{lccccccccc}
  \hline
\bfseries Method&\bfseries AUS&\bfseries BAN&\bfseries BRE&\bfseries CLE&\bfseries DER&\bfseries HOV&\bfseries IRM&\bfseries MAM&\bfseries NTH\\
\hline \hline
\bfseries RNI-II& \bfseries 86.38& \bfseries 69.40& \bfseries 74.83& \bfseries 51.75& \bfseries 95.64& \bfseries 97.02& \bfseries 94.00& \bfseries 83.04& \bfseries 93.92\\
FKMI& 82.03& 68.29& 74.83& 51.45& 95.64& 96.80& 87.33& 82.42& 88.81\\
KMI& 80.29& 68.28& 74.83& 51.43& 95.64&96.80& 92.67& 82.42& 89.74\\
KNNI& 82.90& 69.40& 74.83& 51.75& 95.64& 96.10& 85.33& 82.42& 88.79\\
WKNNI& 80.43& 67.90& 74.83& 51.75& 95.64& 96.10& 85.33& 82.62& 89.26\\
SVMI& 87.25& 69.40& 74.83& 52.44& 95.64& 97.25& 95.33& 83.04& 93.90 \\
\end{tabular}
\end{table}
\begin{table}[!ht]
   \centering
\caption{GA-C4.5 Classifier Test Accuracies}{
    \label{chap9:tab8}
 \begin{tabular}{lccccccccc}
  \hline
\bfseries Method&\bfseries AUS&\bfseries BAN&\bfseries BRE&\bfseries CLE&\bfseries DER&\bfseries HOV&\bfseries IRM&\bfseries MAM&\bfseries NTH\\
\hline \hline
\bfseries RNI-II& \bfseries 85.65& \bfseries 70.14& \bfseries 75.50& \bfseries 52.10& \bfseries 95.09&\bfseries 96.80&\bfseries 94.00& \bfseries 83.35& \bfseries 94.37\\
FKMI& 81.45& 66.80& 75.50& 52.44& 95.09& 96.57& 87.33& 82.42& 88.83\\
KMI& 81.16& 68.65& 75.50& 50.10& 95.09& 96.57& 92.67& 82.42& 88.85\\
KNNI& 81.74& 69.39& 75.50& 50.48& 95.09& 95.88& 85.33& 82.42& 88.81\\
WKNNI& 81.30& 68.46& 75.50& 49.82& 95.09& 95.88& 85.33& 82.52& 89.31\\
SVMI& 85.07& 65.32& 75.50& 49.49& 95.09& 97.25& 95.33& 82.00& 95.30\\
\end{tabular}}
\end{table}

To establish the claim that the improvement in test accuracies obtained by the RNI-II algorithm when compared with other algorithms is not due to sampling errors, we performed the Wilcoxon signed rank test among the matched pairs. The  test results at significance level $\alpha=0.05$ are presented in Table~\ref{chap9:tab9} and Table~\ref{chap9:tab11}.

The Table~\ref{chap9:tab9} depicts the Wilcoxon statistics generated for the test accuracies of C4.5 classifier generated from different data sets which are imputed using RNI-II and other methods. The performance of RNI-II is found to be superior to KMI and FKMI in terms of scoring positive rank sum of $28$ with a p-value of $0.03$. While RNI-II has shown higher positive rank sums with  p-value$\leq0.05$ at a significance level of $0.05$ when compared with other imputation algorithm.

It follows from the Table~\ref{chap9:tab11} the test accuracies of GA-C4.5 have greatly improved when imputation is carried using RNI-II when compared with other imputation methods like FKMI, KMI, WKNNI and KNNI. The RNI-II has scored highest positive rank sum of $28$ in relation to WKNNI, KMI and KNNI with a p-value of $0.01$ at $0.05$ significance level. In the case of SVMI the RNI-II imputation has scored a positive sum of $20$  but could not meet the critical value to attain the required statistical significance level.
\begin{table}[!ht]
\centering
\caption{Wilcoxon sign rank statistics for paired samples Results for C4.5 Classifier}
\label {chap9:tab9}
\begin{tabular}{lcccc}
\hline
\bfseries Method	&	\bfseries Rank Sums &\bfseries Test& \bfseries Critical& \bfseries p-value \\
& \bfseries (+, -)&\bfseries  Statistics&\bfseries Value&\\
\hline \hline
FKMI	&	28.0, 0.0&0.0&3.0&0.02\\
KMI		&28.0, 0.0&0.0&3.0&0.02 \\
KNNI	&	15.0, 0.0&0.0&0.0&0.06\\
WKNNI	&	21.0, 0.0&	0.0&1.0&0.03\\
SVMI	&	1.0,14.0&1.0&0&0.12\\
\end{tabular}
\end{table}
\begin{table}[!ht]
\centering
\quad \quad
\caption{Wilcoxon sign rank statistics for paired samples Results for GA-C4.5 Classifier} \label{chap9:tab11}
\centering
\begin{tabular}{l|c|c|c|c}
\hline
\bfseries Method	&	\bfseries Rank Sums &\bfseries Test&\bfseries Critical&\bfseries p-value \\
& \bfseries (+, -)&\bfseries  Statistics&\bfseries Value&\\
\hline \hline
FKMI	&26.0, 2.0&	2.0&3.0&0.04 \\
KMI		&28.0, 0.0	&0.0	&3.0	&0.02\\
KNNI	&28.0, 0.0	&0.0	&3.0	&0.02\\
WKNNI&28.0, 0.0	&0.0	&3.0	&0.02\\
SVMI&20.0,8.0&8.0&3.0&0.37\\
\end{tabular}
\end{table}

\begin{figure}[!ht]
\centering
  \caption{RNI-II vs FKMI}
  \includegraphics[width=0.5\textwidth]{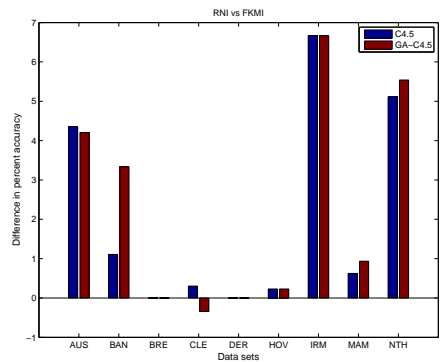}
  \label{chap9:fig1}
  \end{figure}

\begin{figure}[!ht]
\centering
  \caption{RNI-II  Vs KMI}
  \includegraphics[width=0.5\textwidth]{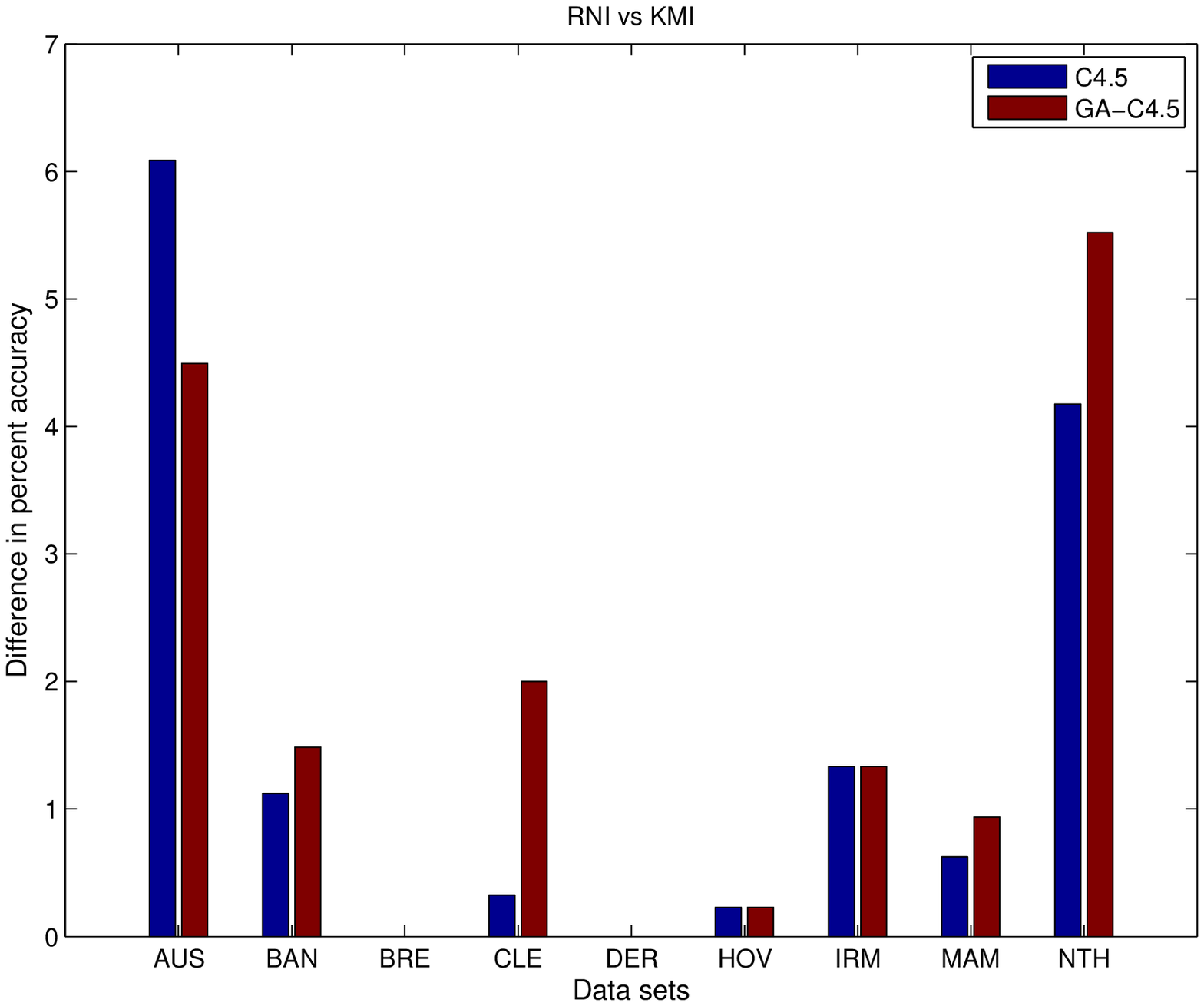}
  \label{chap9:fig2}
  
\end{figure}

\begin{figure}[!ht]
\centering
  \caption{RNI-II  Vs KNNI}
  \includegraphics[width=0.5\textwidth]{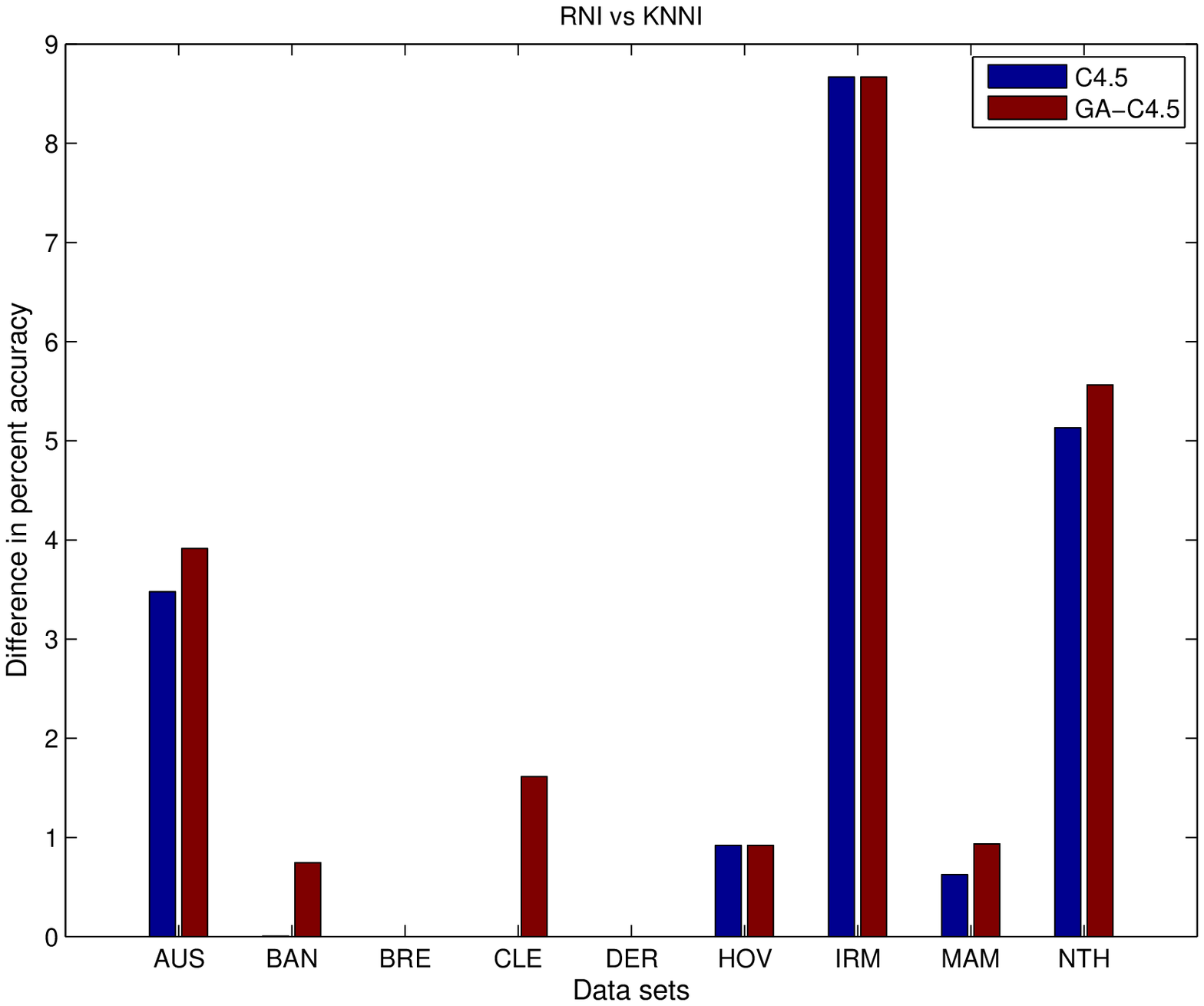}
  \label{chap9:fig3}
  
	\end{figure}

\begin{figure}[!ht]
\centering
  \caption{RNI-II Vs SVMI}
  \includegraphics[width=0.5\textwidth]{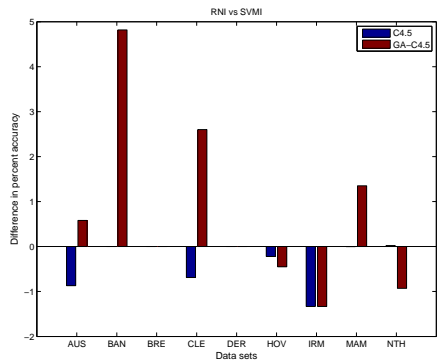}
  \label{chap9:fig4}
  
\end{figure}

\begin{figure}[!ht]
\centering
  \caption{RNI-II Vs WKNNI}
  \includegraphics[width=0.5\textwidth]{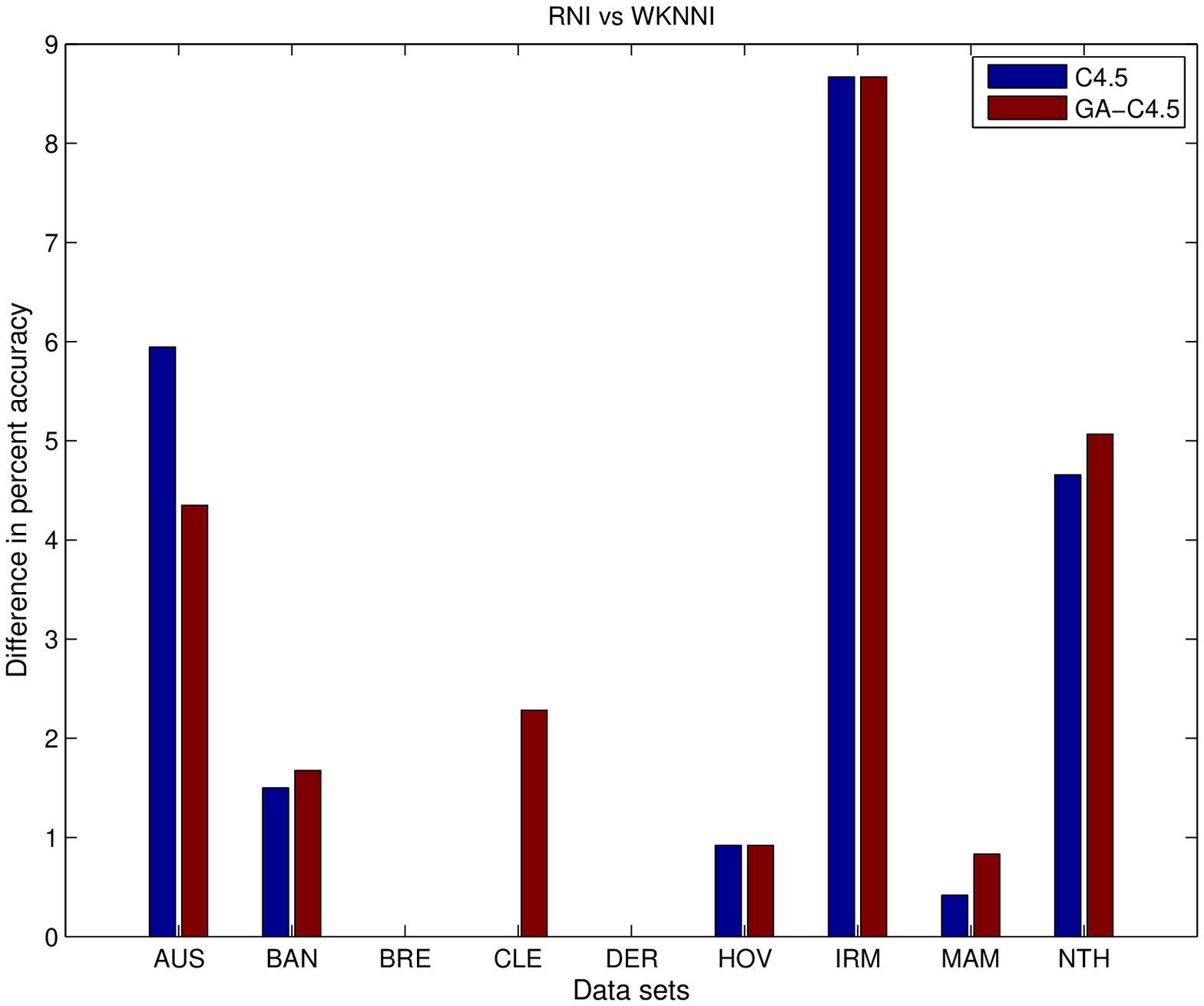}
  \label{chap9:fig5}
  
	\end{figure}

The differences in the accuracies between RNI-II and other imputation algorithms are shown in Figures~\ref{chap9:fig1} to~\ref{chap9:fig5}. In these figures vertical bars represent the difference in test accuracy obtained using RNI-II relative to other imputation methods for a variety of data sets. The positive values (bars above the baseline) represent the improvement in percentage accuracy obtained by using RNI-II method. It may be understood that the larger the bars indicate higher percentage accuracy between RNI-II method with the other base method, while a negative bar (below the baseline) indicates the decision tree classifier has a lower test accuracy for RNI-II relative to other imputation method in question. In case of C4.5 decision tree classifier the RNI-II algorithm has scored around $8\%$ improvement in accuracies as compared to  FKMI, KNNI, WKNNI imputation algorithms for IRM data set.

In the case of GA-C4.5 decision tree classifier we notice that our RNI-II method has achieved highest accuracy in the case of data sets AUS, BAN, and MAM while the performance of other methods (FKMI, KMI, KNNI, WKNNI and SVMI) are far below this accuracy.
\newpage
\section{Conclusions}
\label{chap9:sec6}
In this paper a review on the missing values in databases and possible strategies for imputing missing values is discussed. A mathematical framework for computing an indexing measure is developed which in turn is used in the computation of the distance between the pairs of data records. The RNI-II procedure is presented and performance comparison on benchmark datasets is carried out. The new imputation procedure RNI-II has enhanced the test accuracies of the two decision tree classifiers (C4.5 and GA-C4.5) significantly in comparison to other imputation methods KMI, FKMI, WKNNI and KNNI with a \emph{p} value $<$ 0.05 at significance level $\alpha=0.05$. 

\bibliographystyle{plainnat}

\end{document}